\documentclass{article}
\usepackage{ijcai18}

\usepackage{times}
\usepackage{xcolor}
\usepackage{soul}
\usepackage[utf8]{inputenc}
\usepackage[small]{caption}
\usepackage[normalem]{ulem}

\title{Psychological State in Text: A Limitation of Sentiment Analysis}

\author{
Hwiyeol Jo$^1$, 
Jeong Ryu$^2$, 
\\ 
$^1$ Department of Computer Science \& Engineering, Seoul National University \\
$^2$ Department of Psychology, Seokyeong University \\
hwiyeolj@gmail.com,
ryujeong@gmail.com
}

\begin{document}

\maketitle

\begin{abstract}
  Starting with the idea that sentiment analysis models should be able to predict not only positive or negative but also other psychological states of a person, we implement a sentiment analysis model to investigate the relationship between the model and emotional state. We first examine psychological measurements of 64 participants and ask them to write a book report about a story. After that, we train our sentiment analysis model using crawled movie review data. We finally evaluate participants' writings, using the pretrained model as a concept of transfer learning. The result shows that sentiment analysis model performs good at predicting a score, but the score does not have any correlation with human's self-checked sentiment.
\end{abstract}

\section{Introduction}

Sentiment analysis is a kind of text mining, which is to predict human mind, specifically the emotional state of a person by extracting emotional expressions from text~\cite{pang2008opinion}. Sentiment analysis models have achieved relatively high accuracy despite the ambiguity of both words and emotion itself. One of the commonly used datasets is Stanford Sentiment Treebank (SST), which consists of 5 class labeled reviews (very negative, negative, neutral, positive, and very positive)~\cite{socher2013recursive}. Current state-of-the-art model on SST, Tree-LSTM with refined word embeddings~\cite{yu2017refining} performs 54.0\% on accuracy. On the other hand, as far as we know, there is no research to find that sentiment analysis models could predict emotional state of a person. Starting with the idea that sentiment analysis models should be able to predict not only positive or negative valence but also emotional state, we build a sentiment analysis model to investigate whether the model could predict emotional state or not.

\section{Emotion effects on Memory}

Emotion affects the contents of memories both when we store and retrieve an memory. Theory “Affect as information framework” argues that if we are positive, we tend to be interpretive, relational processing while being detailed, stimulus-bound, referential processing in negative affective~\cite{clore2001affect}. \citeauthor{kensinger2007effects} found different specificity of an event. Emotion can consolidate memory. Emotional memories are kept fairly untouched and slowly forgotten because they are repeatedly rehearsed. Negative emotion can improve an accuracy of the memory~\cite{kensinger2007negative}. Emotion can make affect regulation~\cite{fonagy2018affect,raes2003autobiographical}. If an event has occurred for a long period, people evaluate the event depending on their emotion when emotional arousal is peak and is at the end of the event. Mood-congruent information which is in the same valence with one’s present mood would be more perceived, memorized, retrieved, and utilized to decide than mood-incongruent information. With the backgrounds, we assume that people retrieve memories about the same event differently depending on their emotional state.

\section{Methods}
\subsection{Data Collection}

We collected two different types of data. The first one is experimental data, which include psychological measurements to examine emotional state of participants as well as book reports the participants retrieved and wrote down in the emotional state. The other is movie review data to train sentiment analysis model. The book report data is too few to train the sentiment analysis model, so we have to borrow a concept of transfer learning~\cite{pan2010survey}, which is conveying knowledge from related target task.

\subsubsection{Experimental Data}

All the 64 participants were university students in South Korea, who enrolled in "Introduction to Cognitive Science" class in Konkuk University during 2014 fall semester. Evenly distributed participants were sampled; 39 (61\%) male and 25 (39\%) female and 31 (48\%) students majored Liberal Arts while the others (52\%) majored in Science. Its average and standard deviation of age are 22.5 and 2.42, respectively. However, only 55 completed all the experiments. We examined depression (Center for Epidemiological Studies – Depression)~\cite{radloff1977ces}, present positive affective and negative affective (Positive Affective and Negative Affective Schedule)~\cite{watson1988development} because depression and emotional valence are well-known factors that affect the memory and their retrieval styles~\cite{thomas2007depressed}. CES-D measures how depression is severe during specific periods. PANAS measures positive affectivity (interested, excited, strong, enthusiastic, proud, alert, inspired, determined, attentive, and active) and negative affectivity (distressed, upset, guilty, scared, hostile, irritable, ashamed, nervous, jittery, and afraid).
Participants first completed the psychological measurements. The result is presented in Table~\ref{tab:1}.

\begin{table}[ht] \centering
    \begin{tabular}{|c|c|c|c|c|c|}
    \hline
    Measurement & N & \% & Mean & SD & Range \\
    \hline \hline
    PANAS POS & 55 & 100 & 27.764 & 6.802 & 22-50 \\
    PANAS NEG & 55 & 100 & 34.127 &7.149 & 15-46 \\
    \hline
    PANAS & 55 & 100 & 61.891 & 8.381 & 47-92 \\ 
    \hline \hline
    Depressed & 34 & 61.8 & 33.853 & 8.610 & 21-50 \\
    Non-depressed & 21 & 38.2 & 13.714 & 4.051 & 5-20\\
    \hline
    CES-D & 55 & 100 & 26.164 & 12.202 & 5-50 \\
    \hline
    \end{tabular}
    \caption{The psychological state of participants.}
    \label{tab:1}
\end{table}

Unusually, depressed participants are more than non-depressed. Perhaps, this is because we measured them after a difficult midterm exam and in 2014, South Korea, many people were dead by accidents. After the measurements, we gave them a book, Chronicle of a Death Foretold~\cite{marquez2014chronicle}. To make participants read carefully, we announced that they would take a quiz. After a while, we asked the participants to write book reports with stories of the book as detailed as they could remember. As a result, we could collect 63 book reports in Korean with 133.24 average words where their standard deviation is 52.27. The maximum and minimum number of words in the writings are 20 and 278, respectively.

\subsubsection{Movie Review Data}

We crawled movie review data from Naver, the most famous portal site in Korea. The data consists of the information of titles, scores, and comments on a movie. The distribution of scores in the movie review data is presented in Table~\ref{tab:2}.

\begin{table}[ht] \centering
    \begin{tabular}{|c|c|c||c|c|c|}
    \hline
    Score & N & \% & Score & Number & \% \\
    \hline
    \hline
    1 & 61,307 & 16.76 & 6 & 26 & 7.17 \\
    2 & 8,700 & 2.38 & 7 & 44.736 & 12.22 \\
    3 & 8,674 & 2.37 & 8 & 89,310 & 24.41 \\ 
    4 & 9,223 & 2.52 & 9 & 97,169 & 26.56 \\
    5 & 50,463 & 5.59 & \sout{10} & \sout{327,544} & \\
    \hline
    \end{tabular}
    \caption{The distribution of scores of collected movie review data.}
    \label{tab:2}
\end{table}

However, we found the distribution of scores is biased on 10 points. The reason might be because of fake reviews for advertising, or getting some points by leaving commonplace reviews. The fake data could drive our model to be biased, so we decided to ignore 10 points data with an assumption that 8 or 9 points reviews might include positive expressions as much as 10 points reviews.

\subsection{Sentiment Analysis Model Implementation}

We implement a sentiment analysis model based on well-known deep neural networks, TextCNN~\cite{kim2014convolutional}. The main reasons we choose the algorithm are (i) hard to use state-of-the-art model as it is, because we have different data in a different language. (ii) TextCNN is simple but performs as well as state-of-the-art model. We split our data into train set, validate set, and test set, adapting early-stopping to prevent our model from overfitting. We train the model using the movie review data with the belief that the model will be trained to catch emotional expressions in text. After the training is completed, we evaluate scores of book report data, which is scoring emotional expressions in participants' writing. As a concept of transfer learning, the model for a specific task (scoring the movie review) could be utilized for similar target tasks (scoring the book reports) in the way of extracting expressions and evaluating the expressions in text. This technique helps us to overcome the data problem that only 55 book reports are not enough to train a model as well as experimental data is hard to collect in large scale.

\section{Result}

We first train TextCNN model in collected movie review data and see its performance. On 9-level polarity, TextCNN performs 41.32\% on accuracy. We can ensure that TextCNN performs well on our data in that our data has 9-level polarity whereas state-of-arts model performs 52.0\% on 5-level polarity. After checking the performance, we evaluate book report data; 3 of them are 1 point, 4 of them are 2 points, 2 of them are 3 points, 3 of them are 4 points, 1 of them is 5 points, 29 of them are 8 points. Finally, we investigate the relationship between the scores and psychological measurements. This approach is based on the studies that the content of retrieved memory depends on psychological state~\cite{bower1981mood,blaney1986affect}. The correlations between evaluation scores and psychological measurements are presented in Table~\ref{tab:3}.

\begin{table}[ht] \centering
    \begin{tabular}{|c|c|c|c|c|}
    \hline
    $\rho$ & \small{PANAS POS} & \small{PANAS NEG} & \small{PANAS} & \small{CES-D} \\
    \hline
    \hline
    \small{Score} & 0.030 & 0.096 & 0.108 & 0.255 \\
    \hline
    \end{tabular}
    \caption{Correlations between measurement and model score.}
    \label{tab:3}
\end{table}

Our model scores weakly correlate with only CES-D in positive. However, this means that if a person uses positive words, he/she is depressed. The result is in contradict to general knowledge on emotion. Therefore, we conclude that any of correlation is not meaningful.

\section{Conclusion}

In this study, starting with the idea that sentiment analysis models should be able to predict not only positive or negative valence but also emotional state, we analyze the relationship between scores from a sentiment analysis model and psychological state. We show that sentiment analysis model performs good at predicting a score, but the score does not have any correlation with human's self-checked sentiment.\\
The contribution of this work can be listed as follows: (i) we investigate the relationship between a sentiment analysis model and psychological measurement, claiming that sentiment analysis models should be able to explain not only whether the data contains negative or positive meaning but also whether the person is negative or positive. (ii) we suggest a framework as objective methods to evaluate sentiment analysis model.

\bibliographystyle{named}
\bibliography{ijcai18}

\end{document}